\documentclass[letterpaper, 10 pt, conference]{ieeeconf}

\IEEEoverridecommandlockouts

\usepackage{epsfig}
\usepackage{amsmath}
\usepackage{amssymb}
\usepackage{gensymb}
\usepackage[T1]{fontenc}
\usepackage{algpseudocode}
\usepackage{mathtools}
\usepackage{multirow}
\usepackage{makecell}
\usepackage{bm}
\usepackage{algorithm}

\usepackage{epstopdf}
\usepackage{pifont}
\usepackage{multirow}
\usepackage{url}
\usepackage{tabularx}
\usepackage[linkcolor=black,citecolor=black,urlcolor=black,colorlinks=true]{hyperref}
\usepackage[skip=3pt, font={small}]{caption}
\usepackage{booktabs}

\title{\LARGE \bf
CMoE: Contrastive Mixture of Experts for Motion Control and Terrain Adaptation of Humanoid Robots}

\author{
    Shihao Ma$^{1,*}$, Hongjin Chen$^{1,*}$, Zijun Xu$^{1,*}$, Yi Zhao$^{1}$, Ke Wu$^{1}$, Ruichen Yang$^{1}$,\\
    Leyao Zou$^{1}$, Zhongxue Gan$^{1,\dagger}$, and Wenchao Ding$^{1,\dagger}$
    \thanks{$^{1}$College of Intelligent Robotics and Advanced Manufacturing, Fudan University, Shanghai, China, 200433}
    \thanks{This work was supported in part by the National Natural Science Foundation of China (NSFC) under Grant 62403142, in part by the Science and Technology Commission of Shanghai Municipality under Grant 24511103100, and in part by the Shanghai Municipal Science and Technology Major Project (No. 2021SHZDZX0103).}
    \thanks{$^{*}$Equal contribution. $^{\dagger}$Corresponding authors.}%
\thanks{Project Page: \url{https://hoshi-no-ai.github.io/CMoE}}%
}

\makeatletter
\IEEEaftertitletext{%
  \vspace{-0.1cm}
  \begin{center}
    \includegraphics[width=\linewidth]{figure/teaser.jpg}
    \captionof{figure}{\label{fig:simu_res}We propose a mixture-of-experts model-based architecture that enables humanoid robots to simultaneously navigate a variety of challenging terrains. We validate our strategy on complex mixed terrains and non-training environments.}
    \vspace{0.1cm}
  \end{center}
}
\makeatother

\begin{document}

\maketitle
\thispagestyle{empty}
\pagestyle{empty}

\begin{abstract}

For effective deployment in real-world environments, humanoid robots must autonomously navigate a diverse range of complex terrains with abrupt transitions. While the Vanilla mixture of experts (MoE) framework is theoretically capable of modeling diverse terrain features, in practice, the gating network exhibits nearly uniform expert activations across different terrains, weakening the expert specialization and limiting the model's expressive power. To address this limitation, we introduce CMoE, a novel single-stage reinforcement learning framework that integrates contrastive learning to refine expert activation distributions. By imposing contrastive constraints, CMoE maximizes the consistency of expert activations within the same terrain while minimizing their similarity across different terrains, thereby encouraging experts to specialize in distinct terrain types. We validated our approach on the Unitree G1 humanoid robot through a series of challenging experiments. Results demonstrate that CMoE enables the robot to traverse continuous steps up to 20 cm high and gaps up to 80 cm wide, while achieving robust and natural gait across diverse mixed terrains, surpassing the limits of existing methods. To support further research and foster community development, we release our code publicly.

\end{abstract}

\section{Introduction}
Humanoid robots are expected to operate in abrupt transition real-world environments, where they frequently encounter dynamically changing terrains, such as gravel paths that transition into slopes or stepping stones interspersed with deformable surfaces\cite{Lai2024WorldMP}\cite{romay2014template}. 
Walking in such environments requires not only the ability to maintain stability on a single type of terrain but also the agility to rapidly adapt to heterogeneous terrains while preserving continuous motion.

To enable robots to adapt to various terrains, previous works\cite{robotparkour}\cite{HumanoidParkour} have proposed a two-phase training approach, where the model is first trained on a single terrain and then distilled in a second phase. This method allows the model to focus on the details of each terrain while avoiding catastrophic forgetting that may occur when training on multiple terrains simultaneously. However, although this two-phase training approach enhances multi-terrain learning, it increases the training time and introduces the risk of overfitting.

\begin{figure}[t]
    \centering
    \includegraphics[width=1\linewidth]{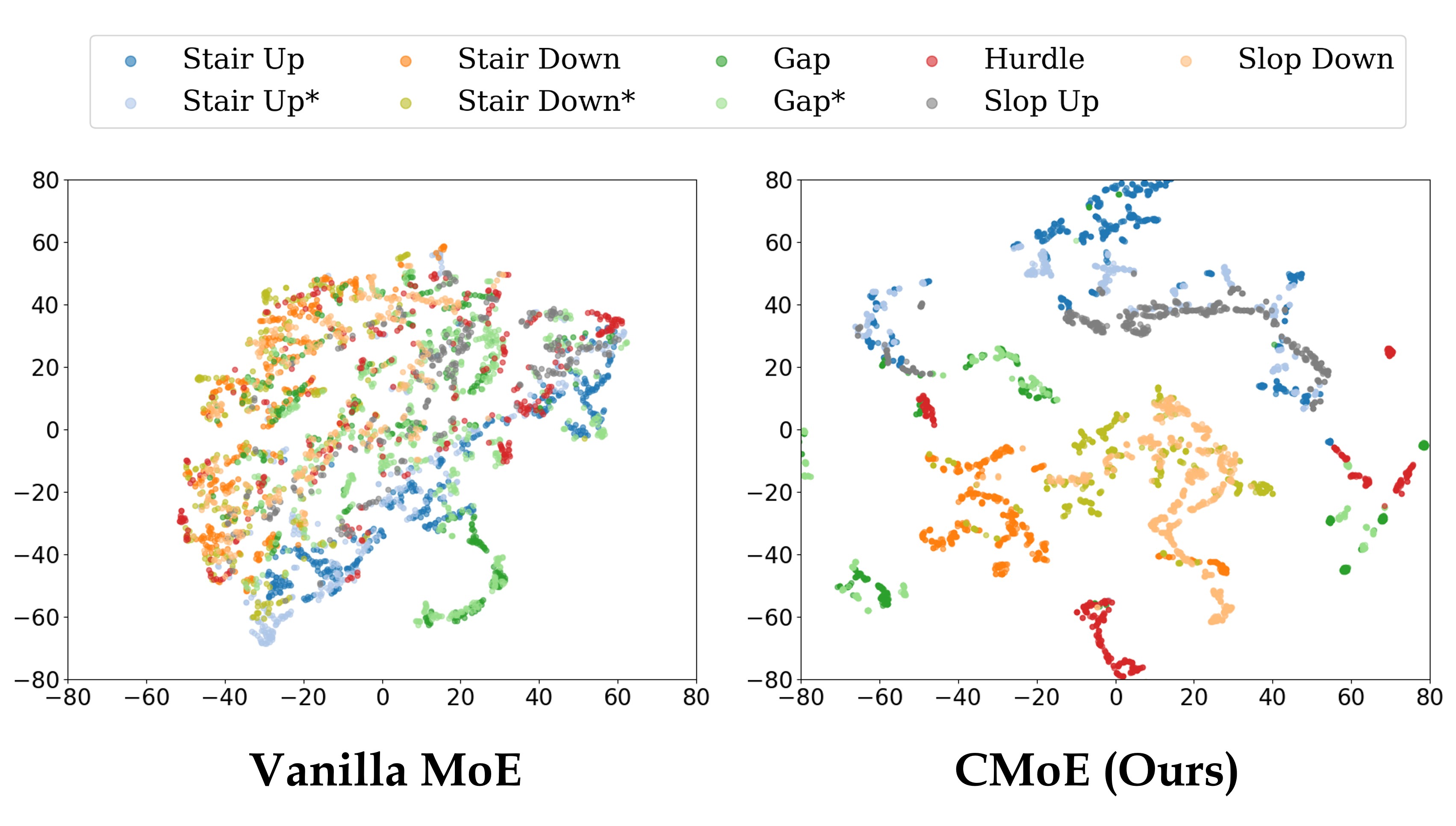}
    \caption{t-SNE visualization of the experts activation of \textbf{Vanilla MoE} and \textbf{our method} across different terrains. ("*" indicates a simple version of this terrain)}
    \vspace{-0.3cm}
    \label{tsne}
\end{figure}

To address these issues, some studies\cite{Huang2025MoELocoMO}\cite{Wang2025MoREMO} have employed the MoE (Mixture of Experts) method, which can simultaneously consider features from multiple terrains. By dynamically activating experts, the MoE network adapts to various terrains and demonstrates good generalization capability. However, we found that it is not effective at activating skills based on environmental features. 
As shown in Fig. \ref{tsne}, the network using the Vanilla MoE strategy exhibits dispersed expert activation across both similar and different terrains, without forming any distinct clusters.
This lack of clear clustering prevents the network from effectively adapting to specific terrain features, leading to less efficient skill activation and terrain handling.

To provide terrain adaptability for the robot, enabling it to effectively activate skills based on environmental features, we propose CMoE, a novel single-stage reinforcement learning framework that integrates the Mixture of Experts (MoE) architecture with contrastive learning. 
Unlike Vanilla MoE, CMoE maximizes the diversity of expert activation distributions within different terrains. Our method effectively activates skills based on environmental features and allocates experts with terrain-specific expertise when transitioning between terrains.

To achieve this, our framework first encodes the perceptual data using an autoencoder (AE), capturing the latent representation of the state of the environment. 
Next, contrastive learning is applied to improve the terrain perception gating network. This method maximizes the consistency of expert activation distributions within the same terrain while minimizing the similarity of expert activation distributions across different terrains. By minimizing the contrastive loss between expert activation levels and terrain information, the gating network is guided to output activations that reflect more terrain-specific differences, thereby encouraging the expert networks to specialize in certain types of terrain. 
Finally, a perceptual gating network dynamically adjusts the activation of experts based on environmental perception. The outputs of each expert are fused according to their activation levels, resulting in an action distribution.

To validate the effectiveness of CMoE, we conducted simulation and field experiments on a Unitree robot. Results show that even when trained simultaneously on eight different environments, CMoE achieves a higher learning ceiling on each terrain. Our robot can use a single policy to traverse obstacles up to 30 cm high, continuous steps up to 20 cm high, and gaps up to 80 cm wide, as shown in the Fig. \ref{fig:simu_res}. Surpassing the most challenging terrains studied in existing research.
Furthermore, CMoE effectively activates specific skills based on environmental characteristics, more efficiently allocating expert activations during terrain transitions. This enables CMoE to excel on mixed terrains.

Summary of our contributions:

\begin{itemize}
    \item We propose a single-stage, end-to-end framework that directly maps multimodal sensory inputs to robot actions, integrating estimators to process both historical information and terrain-specific data.
    \item We introduce a novel MoE actor-critic model, augmented with a contrastive learning objective to enhance the robot’s ability to adapt its terrain response strategy across complex, heterogeneous surfaces. 
    \item Extensive experiments conducted on the Unitree-G1 robot demonstrate our approach's state-of-the-art performance, and we release our code to further contribute to the research community.
\end{itemize}

\begin{figure*}
    \centering
    \includegraphics[width=0.97\linewidth]{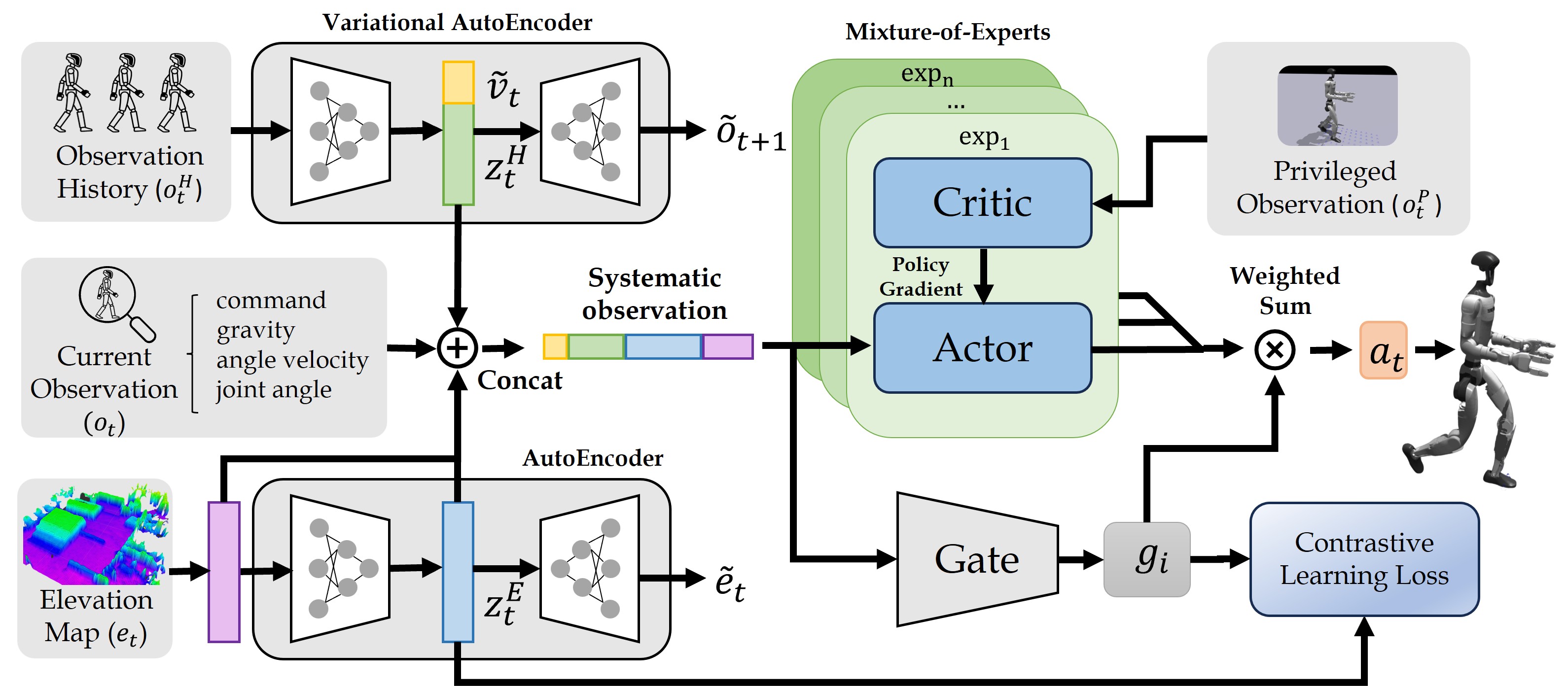}
\caption{Overview of our framework. We encode historical information and the elevation map into explicit and implicit representations, which are fused with the current observation to form the system observation. These are then fed into the MoE structure, consisting of multiple experts and a gating network, which outputs expert activations and performs contrastive learning with the encoded environmental data.}
\vspace{-0.3cm}
\label{pipeline}
\end{figure*}

\section{Related Work}

\subsection{Learning-based Humanoid Locomotion}

With the development of legged robots, especially quadruped robots, motion control methods have matured\cite{Li2024OKAMITH}\cite{Marum2024RevisitingRD}\cite{Zhang2025RENetFM}. Humanoid robots, in particular, have attracted considerable attention due to their ability to better adapt to human-designed environments\cite{Gu2025HumanoidLA}\cite{Gu2024HumanoidGymRL}. However, walking on two legs requires precise center of gravity control and coordination, especially on complex terrain.
To better enable robots to perceive their environment and their own state, some work\cite{vae} uses AE or VAE estimators to encode historical proprioceptive information and employs contrastive loss for state prediction, providing more information. However, due to the lack of environmental sensors, these methods have yet to fully realize the potential of robots.

In response to these limitations, models incorporating external perception have shown promise in enhancing mobility performance. Research has begun using depth cameras or lidar to provide terrain perception information for reinforcement learning models. However, depth camera-based perception strategies are limited by the camera's narrow field of view, resulting in robots being limited to forward movement\cite{Yang2023NeuralVM}\cite{Agarwal2022LeggedLI}\cite{Yu2024WalkingWT}\cite{ExtremeParkour}. LiDAR is widely used due to its wide field of view and higher accuracy. Compared to blind strategies, robots using elevation maps as external perception can better acquire environmental information and make more adaptive actions\cite{PIM}\cite{BeamDojoLA}.

\subsection{Mixture of Experts in Locomotion}
Since the introduction of MoE\cite{Jacobs_Jordan_Nowlan_Hinton_1991}\cite{1994Hierarchical}, it has been regarded as an effective mechanism for addressing gradient conflicts and task interference in multi-task reinforcement learning\cite{Wang2025MultiPLMoEME} due to its modularity and interpretability\cite{Shazeer_Mirhoseini_Maziarz_Davis_Le_Hinton_Dean_2017}\cite{Lepikhin_Lee_Xu_Chen_Firat_Huang_Krikun_Shazeer_Chen_2020}. It has been widely applied in the fields of autonomous driving\cite{Xu2025MoSESM}, natural language processing\cite{Du2021GLaMES}, and computer vision\cite{Huang2024MENTORMN}. 

In recent years, MoE strategies have also been applied in the robotics field, with the introduction of a mixed expert model for quadruped reinforcement learning\cite{Song2024GeRMAG}. The MoE-Loco method\cite{Huang2025MoELocoMO} uses MoE to alleviate gradient conflicts in quadruped and bipedal locomotion tasks, but it lacks an environmental perception system to regulate the MoE. 
MoRE\cite{Wang2025MoREMO} focuses on training a variety of humanoid gaits through MoE, improving the gait learning ability. However, the gait selection is essentially a behavior under human intervention, which does not reflect the robot's autonomy. Moreover, since the expert activation is almost unrelated to the terrain, this does not enhance the robot's terrain traversal capability.
In addition, the current MoE still suffers from the lazy gating problem\cite{Zheng2025RethinkingGM}\cite{Li2023AdaptiveGI}\cite{Liu2022GatingDC}.

\section{Method}

\subsection{System Overview}
As shown in Fig. \ref{pipeline}, CMoE aims to improve the multi-terrain adaptability of humanoid robots by combining contrastive learning and MoE. Specifically, we introduce the CMoE framework, which synergistically integrates three key components: a VAE to infer the robot's true state from incomplete sensory data, an unsupervised contrastive learning mechanism that learns a latent representation to effectively distinguish between terrains, and a MoE architecture that decomposes the complex control task for specialized action generation. By unifying state estimation, terrain perception, and action selection, our framework enables the development of a highly generalized policy, thereby markedly improving the robot's capabilities in multi-terrain locomotion.

\subsection{Problem Description}

We define multi-terrain locomotion as a Markov Decision Process (MDP) with a tuple $\langle S, A, T, R, \gamma \rangle$, where $S$ is the state space, $A$ is the action space, and $s_t \subseteq S$ represents the robot’s state. The system's dynamics are governed by the transition probability $T$, and the agent receives a reward from $R(s, a)$. The discount factor $\gamma \in [0, 1]$ balances immediate and future rewards. We use Proximal Policy Optimization (PPO) to learn the optimal policy $\pi^*$ that maximizes the expected discounted return:
\begin{equation}
\pi^* = \arg \max \mathbb{E}\left[\sum\limits_{t = 0}^\infty \gamma^t R(s_t, a_t)\right].
\end{equation}

\subsection{Information Encoding}
The proprioception observation $o_t$ from sensors is:
\begin{equation}
\mathbf{o_t} = [\omega_t, g_t, c_v^t, \theta_t, \dot{\theta}_t, a_{t-1}], 
\end{equation}
including robot angular velocity $\omega_t$, gravity direction $g_t$, velocity command $c_v^t$, joint angle $\theta_t$, velocity $\dot{\theta}_t$, and the last action $a_{t-1}$. 

We propose a context-state distillation model to capture the robot’s proprioception. The model contains two separate estimators: the first is a VAE-based estimator that predicts the robot’s body state, and the second is used for extracting features from the elevation map.

The encoder maps the input observation $\mathbf{o_t}^H$ to robot body velocity $\textbf{v}t$ and latent representation ${z_t^H}$, which is then decoded to $\tilde{\textbf{o}}{t+1}$.
Specifically, we employ a $\beta$-variational autoencoder ($\beta$-VAE) for the autoencoder setup.
According to \cite{Ji_2022}, estimating body velocity is crucial for terrain traversal. 
The context-state distillation model involves a hybrid loss function:
\begin{equation}
\mathcal{L}_\text{CS} = \text{MSE}(\tilde{\mathbf{v}}_t, \mathbf{v}_t) + \mathcal{L}_\text{VAE},
\end{equation}
where $\text{MSE}(\tilde{\mathbf{v}}_t, \mathbf{v}_t)$ and $\mathcal{L}_\text{VAE}$ represent body velocity loss and VAE reconstruction loss, respectively. $\textbf{v}_t$ is the ground truth given by simulator. The VAE reconstruction loss, $\mathcal{L}_\text{VAE}$ is formulated as:
\begin{equation}
\mathcal{L}_\text{VAE} = \text{MSE}(\tilde{\textbf{o}}_{t+1},\textbf{o}_{t+1}) \\ + \beta D_\text{KL}(q(\mathbf{z}^H_t \mid \textbf{o}^H_t) \parallel p(\mathbf{z}^H_t)),
\end{equation}
here, $\tilde{\textbf{o}}_{t+1}$ is the reconstructed observation at the next time step, $q(\mathbf{z}^H_t \mid \textbf{o}^H_t)$ is the posterior distribution of $\mathbf{z}^H_t$ given $\textbf{o}^H_t$, and $p(\mathbf{z}^H_t)$ is the prior, assumed to be a standard Gaussian distribution.

The other estimator uses an autoencoder to extract features from the elevation map \cite{TerrainMapping} \cite{ElevationMapping} for self-prediction, with the loss function given by:
\begin{equation}
{\mathcal{L}_\text{AE}= \text{MSE}(\tilde{\textbf{e}}_t,\textbf{e}_t),}
\label{vel
}
\end{equation}
where $\textbf{e}_t$ is the ground truth elevation map from the simulator.

\subsection{Mixture of Experts Policy}
Gradient conflicts typically arise in multitask reinforcement learning \cite{yu2019multi}. 
To tackle these challenges, we incorporate a MoE architecture into both the actor and critic networks within the PPO framework. 

Specifically, each expert module comprises a dedicated actor-critic pair, where each critic evaluates only its corresponding actor using privileged observation. Every expert receives the estimated body velocity $\textbf{v}_t^p$, along with implicit contextual state variables $\textbf{z}_t^E$ and $\textbf{z}_t^H$, current observation $\textbf{o}_t^c$ and the elevation map $\textbf{e}_t$, and produces either an action or a value estimate. 
To ensure consistency between policy evaluation and action generation, the same gating network is shared across both the actor and critic MoE components. The final output is obtained as a weighted sum of the expert responses after softmax normalization:

\begin{equation}
\mu_{\text{weighted}} = \sum_{i=1}^{N} \text{softmax}({g}_i) \cdot \mu_i,
\end{equation}
where $\mu_i$ means the $i$-th expert outputs and ${g}_i$ is the expert activation.

\begin{table}
\renewcommand{\arraystretch}{1.2}
\centering
\caption{Terrain description}
\label{tab:terrain_description}
\begin{tabular}{lll} 
\toprule
\textbf{Terrain}         & \textbf{Description}                                                                                              & \textbf{Range}  \\ 
\hline
Slope                    & Slope angle                                                                                                       & 0-20\degree           \\
Stairs                   & Step height~                                                                                                      & 0.05-0.23m     \\
Gap                      & Ditch width~                                                                                                      & 0.1-0.8m       \\
\multirow{2}{*}{Hurdle~} & Hurdle height~                                                                                                    & 0.2-0.4m       \\
                         & Hurdle width                                                                                                      & 0.1-0.3m       \\
Discrete                 & Irregular protrusion height~                                                                                      & 0.1-0.2m       \\ 
\hline
\multirow{2}{*}{Mix1}    & Mixed terrain of gaps and steps,gaps width                                                                        & 0.1-0.8m       \\
                         & Mixed terrain of gaps and steps,steps height                                                                      & 0.1-0.15m      \\ 
\hline
\multirow{3}{*}{Mix2}    & \begin{tabular}[c]{@{}l@{}}Mixed terrain of single-log bridge and steps,\\bridge width\end{tabular}               & 0.5-1.0m       \\
                         & \begin{tabular}[c]{@{}l@{}}Mixed terrain of single-log bridge and steps,\\step height on the bridge~\end{tabular} & 0.1-0.25m      \\
\bottomrule
\vspace{-0.7cm}
\end{tabular}
\end{table}

\begin{table}[b]
    \vspace{-0.3cm}
    \centering
    \caption{Reward Functions}
    \begin{tabular}{lll}
    \toprule[1.0pt]
    \textbf{Term} & \textbf{Equation} & \textbf{Weight} \\
    
    \midrule[0.8pt]
    velocity tracking & $\exp \left\{- {\|\mathbf{v}_{x y}-\mathbf{v}_{x y}^c\|_2^2}/{\sigma}  \right\}$ & $2.0$ \\[0.2ex]
    yaw tracking & $R_{\text{yaw}} = \exp\left(-\left|\psi_{\text{cmd}} - \psi\right|\right)$ & $2.0$ \\[0.2ex]
    z velocity & $\mathbf{v}_z^2$ & $-1.0$ \\ [0.2ex]
    roll-pitch velocity & $\|\boldsymbol{\omega}_{x y}\|_2^2$ & $-0.05$ \\ [0.2ex]
    orientation & $\|\mathbf{g}_{x}\|_2^2 + \|\mathbf{g}_{y}\|_2^2$ & $-2.0$ \\ [0.2ex]
    base height & $\left(h - h^{\text {target}}\right)^2$ & $-15.0$ \\ [0.2ex]
    feet stumble & $\bigvee_{i \in \text{feet}} \left\{ \|\mathbf{F}_{i, xy}\|_2 > 3 \cdot |F_{i, z}| \right\}$ & $-1.0$ \\
    collision & $\sum_{i \in \mathcal{I}_{\text{penalty}}} \mathbb{I}(\|\mathbf{F}_i\|_2 > 0.1)$ & $-15.0$ \\
    feet distance & \makecell[l]{$\min(|p_{y,0} - p_{y,1}| - d_{min}, $ \\ \qquad $d_{max} - d_{min})$} & $0.8$ \\ [0.2ex]
    feet air time & $\sum_{i=1}^2 \left( t_{\text{air}, i} - t_\text{air}^\text{target} \right) \cdot \mathbb{F}_i$ & $1.0$ \\ [0.2ex]
    feet ground parallel & $\sum_{i=1}^2 \text{Var}(\mathbf{p}_{z, i})$ & $-0.02$ \\ [0.2ex]
    hip dof error & $ \sum_{i \in \text{hip joints}} \left| \theta_i - \theta_i^{\mathrm{default}} \right|^2 $ & $-0.5$ \\ [0.2ex]
    dof acc & $ \sum_{i \in \text{all joints}} \ddot{\theta}_i^2 $ & $-2.5e-7$ \\[0.2ex]
    dof vel & $ \sum_{i \in \text{all joints}} \dot{\theta}_i^2 $ & $-5.0e-4$ \\[0.2ex]
    torques & $ \sum_{i \in \text{all joints}} \tau_i^2 $ & $-1.0e-5$ \\[0.2ex]
    action rate & $\|\mathbf{a}_t-\mathbf{a}_{t-1}\|_2^2 $ & $-0.3$ \\[0.2ex]
    dof pos limits & \makecell[l]{$\text{ReLU}({\boldsymbol\theta} - {\boldsymbol\theta}_\text{max}) +$ \\ 
    \quad \quad \quad $\text{ReLU}({\boldsymbol\theta_\text{min}} - {\boldsymbol\theta})$} & $-2.0$ \\[0.2ex]
    dof vel limits & $\text{ReLU}(|\dot{\boldsymbol\theta}| - |\dot{\boldsymbol\theta}_\text{max}|)$ & $-1.0$ \\[0.2ex]
    torque limits & $ \sum_{i \in \text{all joints}} \mathrm{RELU}\left( \hat{\tau}_i - \hat{\tau}_i^{\mathrm{max}} \right) $ & $-1.0$ \\[0.2ex]
    feet edge & $ \mathbf{1}_{\text{foot at edge of the terrain}} $ & $-1.0$ \\[0.2ex]

    \bottomrule[1.0pt]
    \end{tabular}
    \label{tab:reward}
\end{table}

\begin{table*}[htbp]
    \vspace{2mm}
    \centering
    \caption{Quantitative Comparison in Simulation. Metrics Include Success Rate and Average Travel Distance.}
    \vspace{-1mm}
    \label{tab:combined_comparison}
    \setlength\tabcolsep{8pt}
    \fontsize{8}{10}\selectfont
    \begin{tabular}{c|cccccccc}
        \toprule
        \multirow{2}{*}{\textbf{Method}} & \multicolumn{8}{c}{\textbf{Success Rate} } \\
        & \textbf{slope} & \textbf{stair up} & \textbf{stair down} & \textbf{discrete} & \textbf{gap} & \textbf{hurdle} & \textbf{mix1} & \textbf{mix2} \\
        \midrule  
        \textbf{CMoE}   & \textbf{0.991} & \textbf{0.886} & 0.905 & 0.991 & \textbf{0.974} & \textbf{0.987} & \textbf{0.767} & \textbf{0.747} \\
        Vanilla MoE & 0.957 & 0.798 & \textbf{0.908} & 0.987 & 0.818 & 0.970 & 0.605 & 0.662 \\
        Base  & 0.966 & 0.481 & 0.483 & \textbf{1.000} & 0.221 & 0.779 & 0.276 & 0.388 \\
        \midrule
        & \multicolumn{8}{c}{\textbf{Average Travel Distance (m)} } \\
        \midrule
        \textbf{CMoE}   & \textbf{14.870} & \textbf{10.802} & \textbf{10.824} & 13.440 & \textbf{14.876} & 13.470 & \textbf{12.055}  & \textbf{9.750} \\
        Vanilla MoE & 11.675 & 8.898 & 9.250 & 14.870 & 11.980 & \textbf{14.780} & 9.960 & 8.703 \\
        Base  & 12.917 & 8.210 & 8.726 & \textbf{15.280} & 7.385 & 10.124 & 8.209 & 8.848 \\
        \bottomrule
    \end{tabular}
    \vspace{0mm}
\end{table*}

\subsection{Terrains Contrastive Learning}
We introduce a novel method, terrains contrastive learning, in humanoid locomotion. Through this approach, terrains are encoded into latent features and enhance the relation with MoE gate network. Specially, we first adopt two MLP to transform the gate output and elevation map into the same dimension $g^z_t$ and $e^z_t$, respectively. Then, in the contrastive learning process, if a pair of {$\langle g^z, e^z\rangle$ } belong to the same trajectory, they are considered the positive samples. Otherwise, they are negative. The optimization is inspired by SwAV\cite{caron2021unsupervisedlearningvisualfeatures}. To predict the cluster assignment probability $\mathbf{p}^{\text{g}}_t$ and $\mathbf{p}^{\text{e}}_t$ from $g^z_t$ and $e^z_t$. we apply a $\mathcal{L}_2$-normalization on the prototype to obtain normalized matrix $\mathbf{E}=\{ \bar{\mathbf{e}}_1, ..., \bar{\mathbf{e}}_K \}$, and then take a softmax over the dot products of source vectors or target vectors with all the prototypes:
\begin{equation}
\begin{aligned}
    \mathbf{p}^{\text{g}}_t = \frac{\text{exp}(\frac{1}{\tau} {g^z_t}^\top\mathbf{e}_k)}{\sum_{k'}\text{exp}(\frac{1}{\tau} {g^z_t}^\top\mathbf{e}_{k'})},
\end{aligned}
\quad
\begin{aligned}
    \mathbf{p}^{\text{e}}_t = \frac{\text{exp}(\frac{1}{\tau} {e^z_t}^\top\mathbf{e}_k)}{\sum_{k'}\text{exp}(\frac{1}{\tau} {e^z_t}^\top\mathbf{e}_{k'})},
\end{aligned}
\end{equation}
where $\mathbf{p}_t^{\text{g}}$ and $\mathbf{p}_t^{\text{e}}$ are the predicted probability that terrain map to individual cluster with index $k$, while $\tau$ is a temperature parameter.

To obtain $(\mathbf{q}^{\text{g}}_1, . . . , \mathbf{q}^{\text{g}}_K)$ and $(\mathbf{q}^{\text{e}}_1, . . . , \mathbf{q}^{\text{e}}_K)$ for the aforementioned predicted probabilities, while avoiding trivial solutions, the Sinkhorn-Knopp algorithm\cite{knight2008sinkhorn} is applied to both encoders. Now that we have the cluster assignment gate outputs and elevations, the contrastive learning objective is simply to maximize the match accuracy:

\begin{equation}
\mathcal{J}^{\text{SwAV}} = -\frac{1}{2H} \sum^{H}_{t=1} (\mathbf{q}^{\text{g}}_t \log \mathbf{p}^{\text{e}}_t + \mathbf{q}^{\text{e}}_t \log \mathbf{p}^{\text{g}}_t).
\end{equation}

\section{Implementation Details}
\subsection{Training Parameters}
We trained 4096 environments in parallel on IsaacGym using an NVIDIA RTX 4090 computer for 20,000 epochs. We selected 5 experts and used an elevation map covering a 0.7m x 1.1m rectangle around the robot. The parameters used for contrastive learning were as follows: num prototype = 32 and temperature = 0.2, which were chosen to optimize model performance. The terrains used for training in the simulation included eight types of terrain: stairs, gaps, and flat ground, as shown in the table. To reduce the learning difficulty, we employed a curriculum learning mechanism \cite{Rudin2021LearningTW}. Furthermore, since all terrains were trained in the same phase, to balance the difficulty of each terrain, we divided them into simple and complex categories. We then performed velocity command curriculum learning on complex terrains, gradually increasing the magnitude and direction of the velocity commands.

\subsection{Domain Randomization}
To enhance the robot's real-world motion capabilities, we pre-randomized the following parameters during simulation: joint mass and moment of inertia, friction, restitution, motor strength, and motor \( k_p \) and \( k_d \). We also applied a perturbation of up to 30 N to the robot every 16 seconds. For the elevation map, we introduced both delay noise and Gaussian noise, and pre-randomized the offset and rotation of the elevation map. Specifically, we designed nonlinear salt-and-pepper noise to address the unstable extremes that may occur in the elevation map. The update formula for the height points $h(i)$ is as follows.
\begin{equation}
h(i) = 
\begin{cases} 
\mathcal{U}(M, 2M - m) & \text{with probability } p, \\
\mathcal{U}(2m - M, m) & \text{with probability } p, \\
h(i)  & \text{with probability } 1-2p, \\
\end{cases}
\end{equation}
where $M$ and $m$ are the maximum and minimum values at the corresponding height points, $\mathcal{U}(a, b)$ represents drawing from a uniform distribution with the range from $a$ to $b$. In more complex terrain, the salt-and-pepper noise intensity is higher. Furthermore, real-world elevation data is often limited by sensor resolution, exhibiting smooth curved edges rather than sharp corner transitions. To address this issue, we used another domain randomization technique to chamfer the right-angled edges in the elevation map used in the simulator, thereby more realistically simulating the real world. Real world experiments have verified that this method is effective in addressing noise issues in elevation maps.

\subsection{Reward Function}
In addition to referencing existing work and making some minor adjustments, we redefined the foot distance function and implemented penalties for distances that were either too small. We also designed a discontinuous reward that is enabled only under specific terrain conditions. For example, the foot edge reward is activated only when the robot is trained in a hurdle or gap environment. This is because touching the edge of a step or irregular terrain should not be penalized. For a detailed description of the reward function, see TABLE \ref{tab:reward}.

\section{Experiments}

\subsection{Motion Control Performance}
We compare the proposed framework CMoE with the following baselines: (1) Base strategy: The actor-critic network without the MoE structure, but with the same number of parameters as our strategy. (2) Vanilla MoE strategy: Uses the most basic Vanilla MoE structure, but do not include the terrain encoder and contrastive learning. 

We established a benchmark for the robot's motion performance across different tasks. Our benchmark consists of a 3 m $\times $ 18 m runway with different terrains. The terrain details are shown in Table \ref{tab:terrain_description}.
Experimental environment: Each robot was instructed to walk at a speed of 0.8 m/s on a continuous terrain with obstacles for 20 seconds.
If the termination conditions are met, the trial is recorded as a failure, and the environment is reset. Termination conditions include: (1) collision with parts other than the feet, and (2) torso roll or pitch deviation exceeds 1 degree. If the termination conditions are not met and the maximum running time is reached, the trial is recorded as a success.

Record the following indicators:
(1) Success Rate: the proportion of robots that complete the entire journey to the total number of robots in the experiment.
(2) Average Travel Distance: the movement distance in the direction of movement under the travel time limit.

The experimental results shown in Table \ref{tab:combined_comparison} indicate that CMoE outperforms other methods in both success rate and average travel distance, demonstrating strong terrain adaptability. It excels in complex terrains, such as stairs and gaps, achieving higher success rates and longer travel distances. This highlights CMoE's advantage in overcoming gradient conflicts and optimizing motion control. By integrating MoE and contrastive learning, CMoE dynamically adjusts its expert network, enhancing performance in diverse environments, particularly in mixed terrains like mix1 and mix2.

In contrast, Vanilla MoE performs well in some terrains but underperforms in complex ones like gap and mix2, due to the lack of contrastive learning, limiting its adaptability and performance in challenging environments.

\begin{figure}[b]
    \vspace{-0.5cm}
    \centering
    \includegraphics[width=1\linewidth]{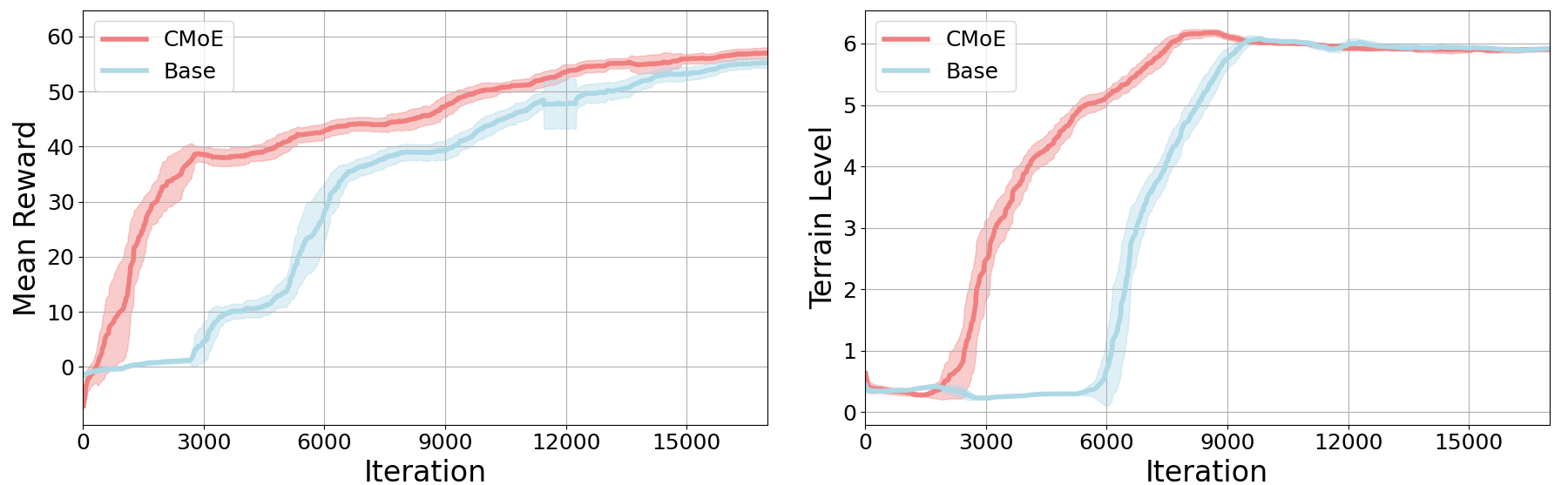}
    \caption{The training curve during the multi-terrain training phase, depicting the reward curve and terrain level change with training iterations.}
    \label{reward}
\end{figure}

\subsection{Effect of Contrastive Learning}
To explore how MoE allocates experts according to terrain and to control for variables, we compared our method (CMoE) with the Vanilla MoE model that does not use contrastive learning, as shown in the t-SNE plot in Fig. \ref{tsne}.

The t-SNE plot of our method shows that similar terrains are clearly clustered based on their degree of similarity, while dissimilar terrains exhibit distinct boundaries. Similar terrains can be grouped and stratified according to their similarity. For example, ascending steps, simple steps, and uphill terrain are clustered together, but also stratified based on terrain difficulty.
Furthermore, we observe that in the t-SNE plot of our method, ascending steps are significantly farther away from descending steps. This suggests that the model does not simply categorize “steps” as one terrain type, but instead differentiates between “ascending steps” and “descending steps,” which aligns with human cognition of walking.

In contrast, the expert weights in the Vanilla MoE model do not show significant changes with different terrains. This indicates that the Vanilla MoE is unable to understand the specialization differences between terrains, limiting the experts’ ability to specialize in different environments. In contrast, CMoE effectively allocates experts according to terrain, as demonstrated by the distinct clustering of expert weights in the t-SNE plot.

\subsection{Expert Behavior Analysis}
\begin{figure}
    \centering
    \includegraphics[width=1\linewidth]{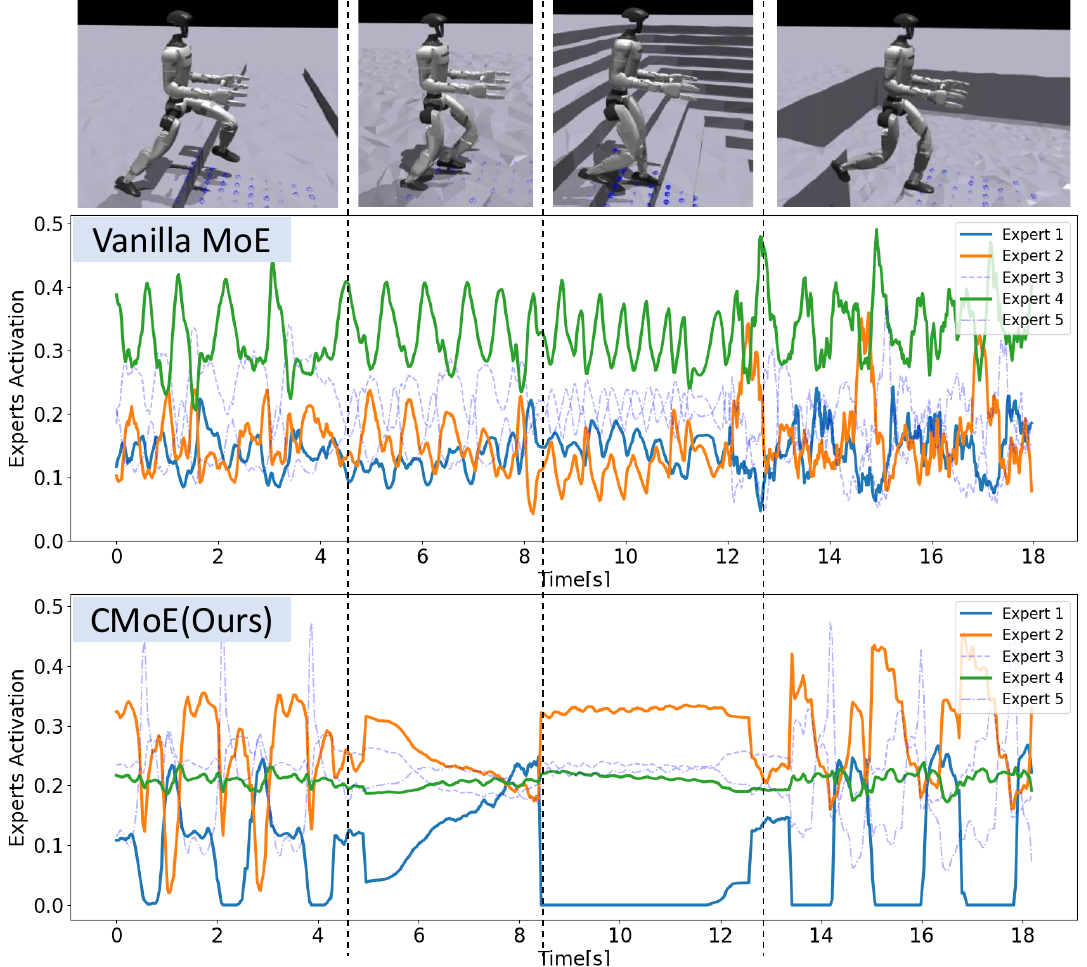}
    \caption{The robot passes through four types of terrain: hurdles, uphill, downstairs, and gaps. The Vanilla MoE method and our method show the changes in the expert activation level over time.}
    \label{fig:placeholder3}
    \vspace{-0.5cm}
\end{figure}

To visually demonstrate the contribution of MoE to the expert activation levels across different terrains, as shown in Fig. \ref{fig:placeholder3}, we designed four different terrains, including three consecutive 30cm hurdles, a 15-degree uphill slope, 10 steps downhill, and three 60cm gaps. We then had robots using the Vanilla MoE and CMoE methods walk across this mixed terrain, recording the changes in expert activation levels as the terrain varied.

It can be observed that in Vanilla MoE, the activation level of each expert remains within a small fixed range with minimal fluctuations, and there is no noticeable change as the terrain changes. In contrast, in CMoE, each expert's activation level exhibits jumps and a wider range of variations, with clear changes occurring as the terrain switches.

Additionally, in the CMoE plot, we can see that Expert 1 only appears in locations where the terrain ascends, such as uphill slopes and hurdles, which require foot lifting. On the other hand, Expert 2 performs oppositely, with a higher weight in areas where the terrain descends. 

We hypothesize that Expert 1 specializes in terrains such as ascending steps. To verify this hypothesis, we removed Expert 1's output from the evaluation environment, while keeping the outputs of other experts unchanged. The experimental results showed that the robot attempted to lift its leg before stepping up the stairs but failed, causing it to stumble and fall. However, in the downhill scenario, the robot successfully navigated the stairs. This indicates that while the robot failed to lift its legs properly, its other movements were unaffected, allowing it to navigate terrains that do not require leg lifting, such as downhill slopes and descending stairs. 
This indicates that Expert 1's specialization is indeed in ascending step terrains, making it the expert for handling such terrains.

\subsection{Training Performance}
To test our method's improvements in training speed and efficiency, we trained using both the Base method and our own method with curriculum learning, recording changes in terrain level and reward.
As shown in Fig. \ref{reward}, the CMoE method’s terrain level increased first and reached its maximum value, indicating that it learned to navigate the most difficult terrain first. The CMoE method outperformed the Base method in both reward growth rate and maximum value, suggesting that the gating network's terrain classification reduces learning difficulty and improves efficiency.

\subsection{Real-world Experiment}
To validate the effectiveness of our approach, we deployed the trained CMoE policy on a Unitree G1 humanoid robot. We used radar to collect point cloud information and combined it with the robot's positioning system to obtain elevation map observations. This data was then packaged and sent to our network, allowing the policy to be directly transferred from the simulated environment to the robot.

We then conducted experiments on various terrains to test the robot's real-world performance. For gap terrain, we tested the robot's maximum traversable width of 80 cm, which, to our knowledge, is the largest among existing methods. For step terrain, we prepared three different step heights: 12, 15, and 20 cm. Our strategy enabled the robot to successfully traverse the most challenging steps, surpassing existing methods in the literature \cite{HIM} (which only support a maximum height of 15 cm). For hurdle terrain, the robot was able to cross thresholds of 30 cm in height. Finally, the robot can easily go up and down a 17-degree slope.

\begin{figure}
    \centering
    \includegraphics[width=1\linewidth]{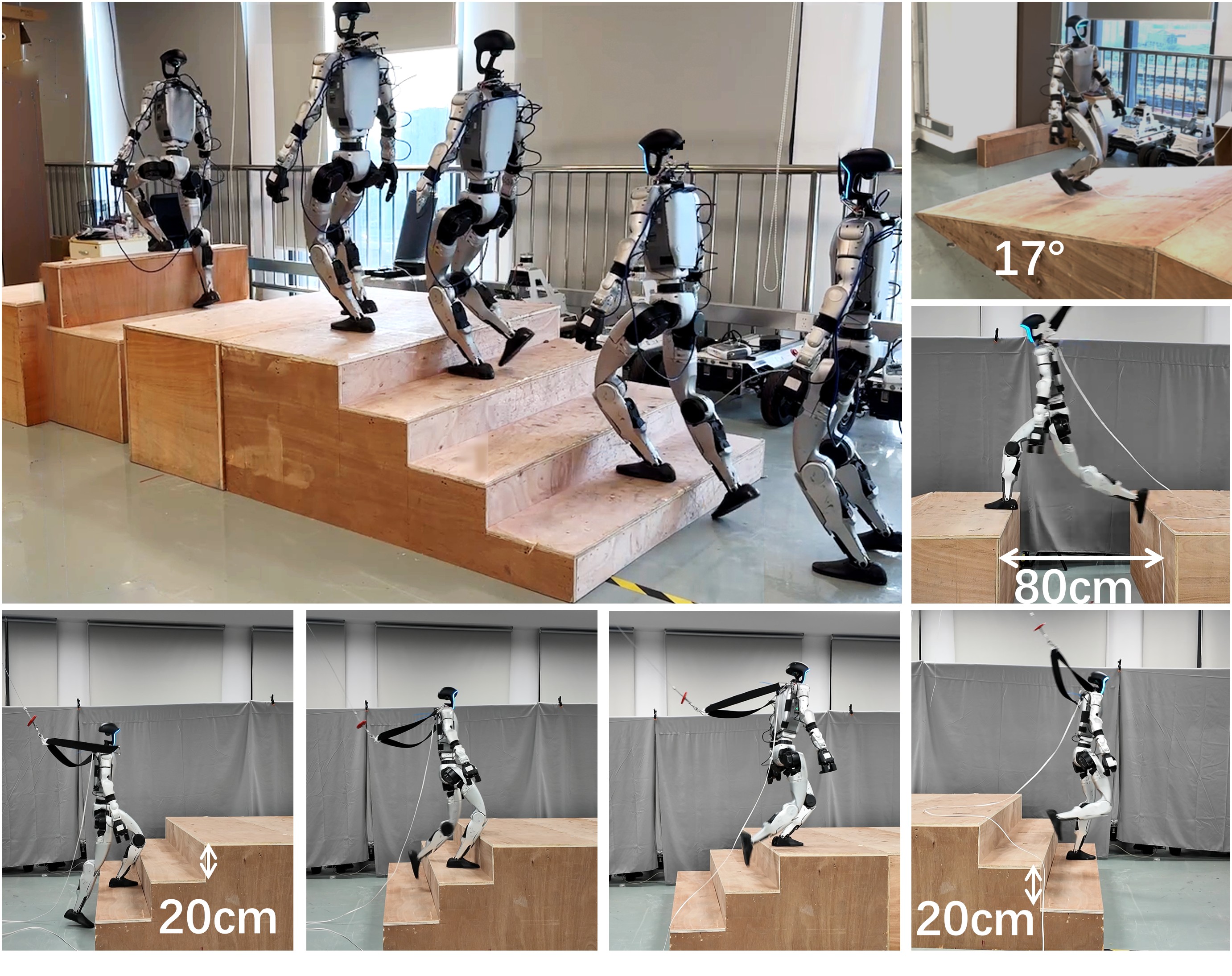}
    \caption{In mixed terrain, the robot climbed 15 cm steps, a 60 cm gap, a 30 cm hurdle, and an uphill slope. In single terrain testing, the robot was able to ascend three steps of 20 cm in height with a steady pace, as well as descend stairs.}
    \vspace{-0.3cm}
    \label{fig:placeholder}
\end{figure}

In addition to these basic terrains, we also tested the robot’s performance on a mixed terrain, which consisted of steps, gaps, slopes, and hurdles. This more complex terrain significantly challenged the robot's ability to adapt to rapidly changing environments. As shown in Fig. \ref{fig:placeholder} and Fig. \ref{fig:simu_res}, the Unitree G1 robot successfully completed a series of terrain crossings, including combinations of the aforementioned single terrains, demonstrating the effectiveness of our method in perceiving various terrains and executing extreme parkour maneuvers.

\begin{figure}
    \centering
    \includegraphics[width=1\linewidth]{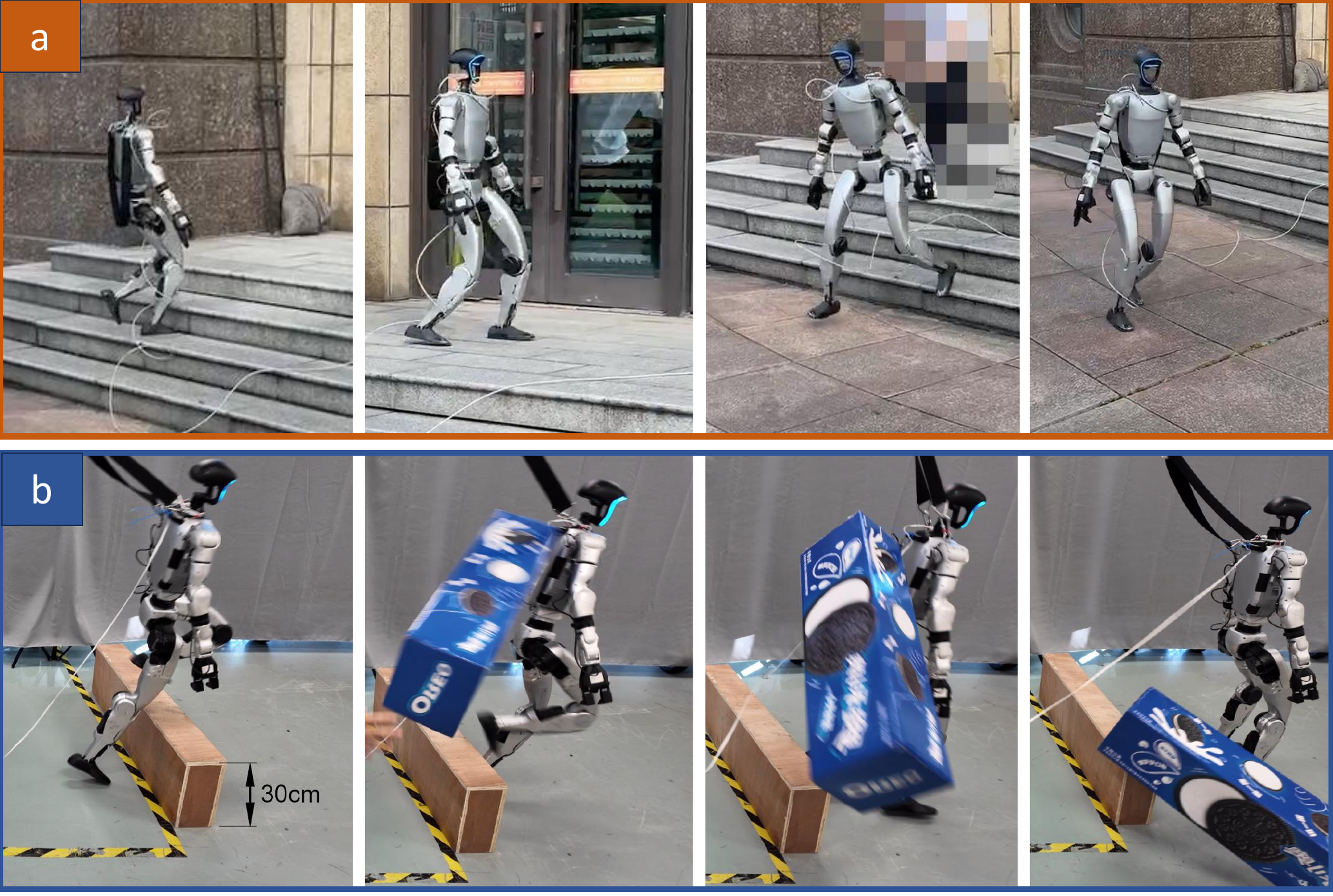}
    \caption{(a) In an outdoor environment, the robot can navigate four continuous sections of untrained terrain with step edges while maintaining continuous movement. (b) Even when encountering human intervention while crossing a 30 cm high obstacle, it can maintain a stable posture and successfully pass the obstacle.}
    \vspace{-0.7cm}
    \label{fig:placeholder2}
\end{figure}

Finally, we conducted robustness experiments to further evaluate the robot’s performance, as shown in Fig. \ref{fig:placeholder2}(a). The step structure in this environment differs in width and height from the stairs in the training environment. Some steps even have protruding edges, which impose higher demands on the robot’s stair-climbing gait. Despite these challenges, the robot still demonstrated excellent stability during long-distance walking and stair navigation.
Additionally, we tested the robustness of our method against disturbances, such as rope dragging and object collisions. As shown in Fig. \ref{fig:placeholder2}(b), when the robot was crossing a 30 cm obstacle and was struck by an object during interference, it remained stable and successfully passed through. These results showcase the strong robustness of our method.

\section{Conclusions}
We propose the CMoE, which integrates MoE and contrastive learning to enhance the robot’s adaptability across diverse terrains. Through a series of experiments, we have demonstrated the superior performance of CMoE in various terrains, including improvements in success rate, travel distance, and robustness in complex environments. Compared to the traditional Vanilla MoE method, CMoE more effectively allocates the expert network, particularly in handling complex terrains, and shows greater stability and generalization ability. The experimental results highlight CMoE's significant advantages in enhancing the robot's control and adaptability, enabling it to handle rapidly changing environments and perform challenging tasks. Future work will extend this approach to whole body control, enabling coordinated full-body parkour.

\bibliographystyle{unsrt}
\bibliography{ref}

\end{document}